\documentclass[runningheads]{llncs}
\usepackage[T1]{fontenc}
\usepackage{lmodern}
\usepackage{siunitx}
\usepackage{booktabs}
\usepackage{tabularx}
\usepackage{multirow}
\usepackage{svg}
\usepackage{graphicx}
\usepackage{mathtools}
\usepackage{amsmath}
\usepackage{amssymb}
\newcommand{\N}{\mathbb{N}}
\newcommand{\R}{\mathbb{R}}
\newcommand{\Z}{\mathbb{Z}}
\newcommand{\abs}[1]{\left\lvert#1\right\rvert}
\begin{document}
\title{$\text{Log}_\text{b}$Quant: Quantizing Language Models in Logarithmic Space}
\author{Jeremias Bohn \and
Tizian Dippold \and
Mahdi Koubaa \and
Elias R.\ Wahl \and
Georg Groh}
\authorrunning{J. Bohn et al.}
\institute{School of Computation, Information and Technology, Technical University of Munich, Munich, Germany\\
\email{\{firstname.lastname\}@tum.de}}
\maketitle              % typeset the header of the contribution
\begin{abstract}
Quantization has become an invaluable tool to reduce memory requirements and inference speed of modern language models, in particular to make them available for consumer setups and edge devices.
While previous work has primarily focused on uniform quantization codebooks, such approaches are prone to suboptimal representations due to low-frequency high-magnitude weights.
We introduce Log$_\text{b}$Quant, a novel logarithmic quantization approach with adjustable bases, to adapt to common parameter distributions.
We show that our method exhibits superior performance at 4-bit precision on several performance benchmarks compared to asymmetric linear quantization at tensor-wise granularity,
while achieving moderate speedup and high memory savings, making it suitable for private use on consumer-grade GPUs.
\end{abstract}
\section{Introduction}\label{sec:introduction}
While large language models and their applications have reached the mainstream over the last few years, most models are closed source and only accessible via API and web interfaces.
However, even open source models, such as Meta's Llama~\cite{grattafiori2024llama3herdmodels} and Qwen's~\cite{Qwen32025} models, come with hardware requirements that often exceed common private setups, which usually comprise a single consumer-grade GPU or edge devices like smartphones.
A significant bottleneck in such setups is the memory size, acting as a hard entrance barrier for local model inference.
This is further exacerbated by the memory's speed, which acts as a primary latency source for small batch sizes (which are common in a local scenario), also known as the \emph{memory wall}~\cite{gholami2024aimemorywall}.

While several approaches exist to reduce parameter count or improve caching, which also reduce memory load and bandwidth utilization, our work focuses on weight quantization, which provides a quick and performance-efficient measure for end users to tackle these issues.
Though most quantization approaches rely on asymmetric linear codebooks, we show that quantization in logarithmic space better retains performance across several benchmarks, in particular in the low-bitwidth regime.
\begin{figure}
  \centering
  \includegraphics[width=0.75\textwidth]{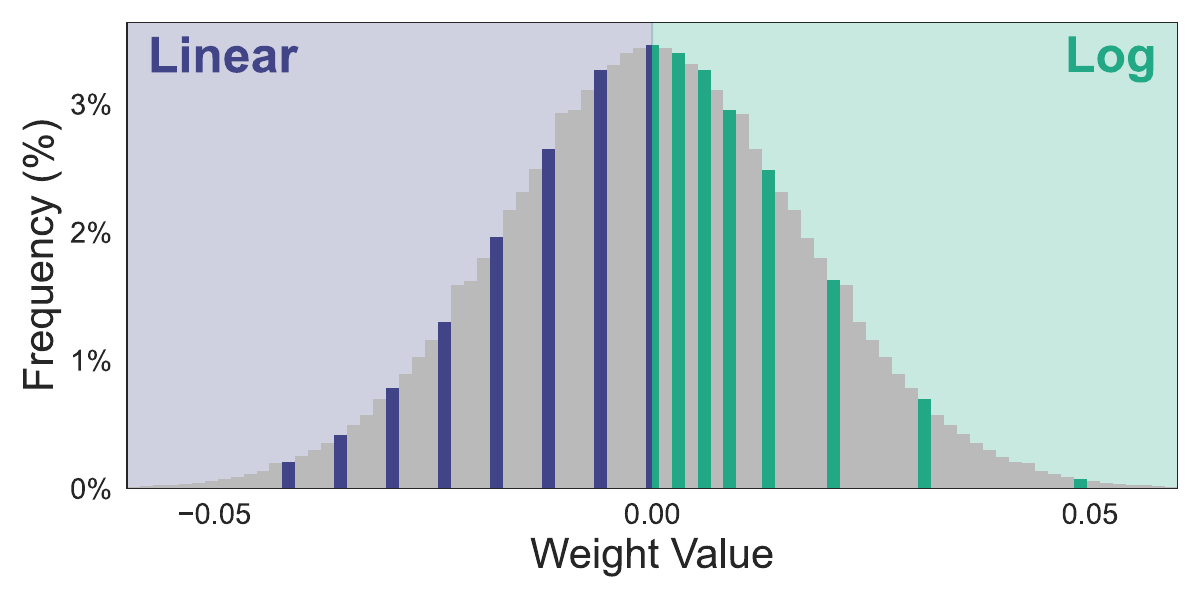}
  \caption{Weight distribution of Llama-3.1-1B's first layer up-projection. Left side: Negative half of 4-bit linear codebook marked in blue. Right side: Positive half of 4-bit Log$_\text{b}$Quant codebook marked in green.}
  \label{fig:weight_dist}
\end{figure}
\section{Related Work}\label{sec:related-work}
Approaches to reduce compute and memory requirements for machine learning models were already discussed early on~\cite{lecun1989optimal,hassibi1992second}, since accelerating neural networks has been and still is a significant issue.
While several approaches consider pruning~\cite{sun2024simple,ma2023llm,ling2024slimgpt} or model compression~\cite{bohn2025adaptive,hsulanguage,lin2025modegpt}, quantization approaches, which reduce representation precision of weights and/or activations, faced widespread adoption due to their easy application and low impact on model performance.

One commonly distinguishes between \emph{uniform quantization}~\cite{gholami_survey_2021}, i.e., using equidistant codebook entries, and \emph{non-uniform quantization}, such as our own approach, Log$_\text{b}$Quant.
Uniform quantization can be described as a linear function $\mathrm{Quant}(s,z,x)=\lfloor \frac{x}{s}\rceil - z$ with scaling factor $s$ and zero-point $z$ (where $\lfloor\cdot\rceil$ is a rounding function).
Such approaches are favorable due to their arithmetic properties, but may spend too much precision to accommodate for low-frequency values (see Figure~\ref{fig:weight_dist}, left side).
On the other hand, non-uniform quantization better adapts to the value distribution of modern neural networks, which commonly are densely centered around zero with few outliers~\cite{miyashita_convolutional_2016,pospieszny_adalog_2025,gholami_survey_2021,vogel_efficient_2018,geng_lookup_2025}.
Previous work which, as our own method, applies a logarithmic scale codebook mostly focused on predefined bases~\cite{miyashita_convolutional_2016,berger2022,Cai2018,pospieszny_adalog_2025,przewlocka-rus_power--two_2022}, mostly base 2 or powers thereof, which may not optimally fit distributions.
On the other hand, we consider bases (and ranges) as optimizable parameters for each layer to fit our codebook to the weight distribution.

Quantization methods further differ in scope: \emph{Weight-only quantization} reduces model parameter precision while activations are kept at full precision to reduce memory footprint and bandwidth usage~\cite{gholami_survey_2021,frantar2023gptqaccurateposttrainingquantization}.
\emph{Activation quantization} quantizes hidden representations~\cite{xiao2024smoothquantaccurateefficientposttraining}, which becomes particularly relevant for the now common GLU-based MLP~\cite{shazeer2020glu} models, which project into a high dimensional space.
\emph{Weight-and-activation quantization} apply precision reduction to both, which offers superior compression and speedups, but also introduces larger algorithmic and numerical challenges~\cite{bondarenko2021understandingovercomingchallengesefficient}.
As our focus lies on consumer and edge devices, where decoding of low batch sizes is memory-bound, we focus on a weight-only approach which comes with significant memory savings and preserves performance more reliably than more aggressive weight-and-activation quantization~\cite{gholami2024aimemorywall,lin2024awqactivationawareweightquantization,frantar2023gptqaccurateposttrainingquantization}.

Lastly, to recover from quantization errors, two major calibration techniques are applied.
In Quantization-Aware Training (QAT), the model is trained to unlearn the precision loss using gradient-based optimization.
QAT approaches, such as Differentiable Soft Quantization~\cite{gong_differentiable_2019} often re-enables superior performance, but requires extensive training and data and faces issues with non-differentiable rounding operations in quantization functions~\cite{yin2019understanding}.
Due to these issues, Post-Training Quantization (PTQ) methods like VPTQ~\cite{liu2024vptq}, GPTQ~\cite{frantar2023gptqaccurateposttrainingquantization}, and GPTAQ~\cite{li2025gptaq} are often preferred as those can quantize models with a small calibration dataset or completely data-free~\cite{nagel2019data} quickly.
In our work, we use GPTQ due to its widespread use and easy adaptability to new quantization functions.
\section{Method}\label{sec:method}
We introduce Log$_\text{b}$Quant, a novel symmetric quantization scheme in logarithmic space.

Our approach is based on the observation that weights in language models are approximately normal distributed (see Figure~\ref{fig:weight_dist}), therefore we shift precision towards distributionally dense areas.
Log$_\text{b}$Quant reduces the need to operate on higher granularity levels, and instead allows tensor-level quantization with minimal performance loss.
\subsection{Quantization Scheme}\label{subsec:scheme}
As explained above, balancing the number of weights mapped to each entry of our codebook $C$ is central to our approach.
We therefore define one codebook per weight matrix, parameterized by base $b\in\R$ and shift $s\in\R$.
Consider any weight $x\in\R$, any quantized weight $q\in\Z$ and any codebook bitwidth $n\in\N$.
Then we define our quantization and dequantization function as
\begin{align}
    \mathrm{Quant}(b,s,x) &\coloneqq
    \begin{cases}
        0 & x=0\label{eq:quant}\\
        \mathrm{sign}(x)\left\lfloor \log_b(\abs{x})+s\right\rceil_{-q_\mathrm{max}}^{q_\mathrm{max}} &x \neq 0
    \end{cases}\\
    \mathrm{Dequant}(b,s,q) &\coloneqq
    \begin{cases}
        0 & q=0\\
        \mathrm{sign}(q)\cdot b^{\abs{q}-s} &q \neq 0
    \end{cases}\label{eq:dequant}
\end{align}
where $q_\mathrm{max}\coloneqq 2^{n-1}-1$ and $\lfloor\cdot\rceil_a^b$ is rounding to the nearest integer with clamping.
Given an effective weight range $[-w_\mathrm{min}, -w_\mathrm{max}]\cup[w_\mathrm{min}, w_\mathrm{max}]$ of a single matrix, we can define $b$ as
\begin{equation}
    b\coloneqq\left(\frac{w_\mathrm{max}}{w_\mathrm{min}} \right)^{\frac{1}{q_\mathrm{max}}}
\end{equation}
which ensures $q_\mathrm{max}$ logarithmic steps from $w_\mathrm{min}$ to $w_\mathrm{max}$.
To align the value $w_\mathrm{max}$ with $q_\mathrm{max}$ in the codebook, we eventually define $s$ as
\begin{equation}
    s\coloneqq q_\mathrm{max} - \left\lfloor \log_b w_\mathrm{max} \right\rfloor
\end{equation}
While these values are not necessarily optimal in terms of codebook balancing, they act as a heuristic to approximate the dynamic range of the given matrix without the need of a non-convex optimization of Equation~\ref{eq:quant}.
\subsection{Energy-Based Pruning}
As seen in Figure~\ref{fig:weight_dist}, weight distribution density is generally highest around $0$.
However, previous works on model pruning~\cite{hassibi1992second,Han2015,cheng2024survey} already established that weights of low magnitude generally contribute less to a model's calculations and can be set to $0$ with low impact on predictions.
Even though exploiting unstructured sparsity is usually difficult, we can use such an approach to shift our codebook density to those parts where it actually matters.
We follow a signal-energy-based pruning strategy and use this to define the effective weight range in Subsection~\ref{subsec:scheme}.
While we can define the maximum in a straight-forward manner as $w_\mathrm{max}\coloneqq \max_{i} \abs{w_i}$, we prune weights by defining our minimum as:
\begin{equation}
    w_\mathrm{min}\coloneqq \min\left\{ t \mid \frac{\sum_{\abs{w_i}\le t} w_i^2}{\sum_i w_i^2}\ge \varepsilon \right\}
\end{equation}
This allows us to discard a fraction $\varepsilon$ of the total energy.
The approach is self-calibrating across layers: In distributions with low standard deviation, the weights close to $0$ hold a significant part of the total energy, thus reducing $w_\mathrm{min}$, while matrices with flatter weight distribution remove larger weights.
We conducted a hyperparameter search, which revealed that generally $\varepsilon=4\times 10^{-3}$ and $\varepsilon=1\times 10^{-6}$ work best for 4-bit and 8-bit quantization, respectively.
\subsection{Implementation}
Since we use a weight-only quantization approach, we have to dequantize the weights at inference to calculate linear layer outputs.
However, as the exponentiation of floating points in Equation~\ref{eq:dequant} for every weight is computationally expensive, we instead rely on a lookup table (LUT) to retrieve dequantized weights directly.
To this end we employ the FLUTE kernel~\cite{bruggemann_flute_2023}, which also directly allows us to apply weight packing for 4-bit quantization.
As FLUTE is designed for blockwise quantization approaches, we simulate our tensor-wise method using a group size of 64, which we chose after a hyperparameter search optimizing for speed benchmarks.
\section{Experiments}
To assess quality and efficiency of our quantization scheme, we evaluate our approach across 8 different text-only decoder-based language models from two model families: Qwen3 and Llama-3.1/3.2.
We compare the models' performance in terms of downstream task accuracy and perplexity and its efficiency in terms of tokens throughput and peak memory usage against a standard asymmetric linear quantization scheme following the methodology of~\cite{dettmers2022llmint88bitmatrixmultiplication}.
\subsection{Setup}
Throughout all our experiments we use the post-training quantization approach GPTQ~\cite{frantar2023gptqaccurateposttrainingquantization} for model calibration, implemented as a Hugging Face Transformers~\cite{wolf-etal-2020-transformers} pipeline.
For our calibration data, we follow the methodology described in GPTQ by randomly (using seed 0) sampling 128 samples of 2048 tokens from WikiText-2's~\cite{merity2016pointersentinelmixturemodels} train split.

We apply a tensor-based weight-only quantization approach both for the asymmetric linear and our own logarithmic scheme, which we apply to the models Llama-3.1-8B, Llama-3.2-1B and -3B, as well as the Qwen3 models of size 0.6B, 1.7B, 4B, 8B, and 14B.
In all cases, the embedding layer as well as the language modeling head are kept in full 16-bit precision (bfloat16), while the remaining weights were reduced to 4- or 8-bit (int8).

For performance benchmarking, we use Eleuther AI's LM Evaluation Harness~\cite{eval-harness}.
Efficiency benchmarking is conducted using PyTorch's~\cite{ansel_pytorch_2024} benchmark framework.

All experiments were carried out on a single Nvidia A40 GPU with 48GB VRAM, alongside an AMD Epyc 7413 24-core CPU @2.65GHz and 512 GB DDR4 RAM\@.
\subsection{Performance Benchmarks}
Downstream performance is measured in terms of accuracy on MMLU~\cite{hendrycks2021measuringmassivemultitasklanguage} (5-shot) and normalized accuracy on PIQA~\cite{bisk2019piqareasoningphysicalcommonsense} and ARC-C~\cite{clark2018thinksolvedquestionanswering} (both zero shot), respectively.
The results can be found in Table~\ref{tab:accuracy}.
Performance of the unquantized models can be found in Table~\ref{tab:accuracy_base} (Appendix~\ref{ap:tabs}).
\begin{table*}[h]
\centering
\setlength{\tabcolsep}{4pt}

\caption{Accuracy ($\uparrow$) on MMLU and normalized accuracy ($\uparrow$) on ARC-C and PIQA, respectively. We compare asymmetric linear quantization vs.\ our Log$_\text{b}$Quant approach.}
\label{tab:accuracy}
\scriptsize
\begin{tabularx}{\textwidth}{l!{\hspace{2pt}\vrule width 0.8pt\hspace{2pt}}l*{8}{>{\centering\arraybackslash}X}}
\toprule

\multicolumn{2}{c}{\multirow{3}{*}{Model}} &
\multicolumn{2}{c}{MMLU} &
\multicolumn{2}{c}{ARC-C} &
\multicolumn{2}{c}{PIQA} &
\multicolumn{2}{c}{Avg.} \\

\cmidrule(lr){3-4}
\cmidrule(lr){5-6}
\cmidrule(lr){7-8}
\cmidrule(lr){9-10}

\multicolumn{2}{c}{} &
lin & log
& lin & log
& lin & log
& lin & log \\
\midrule

\multirow{8}{*}{\textbf{8-bit}}
& Llama-3.2-1B & \textbf{28.79} & 26.32 & \textbf{37.12} & 33.96 & \textbf{73.94} & 73.56 & \textbf{46.62} & 44.61 \\
& Llama-3.2-3B & \textbf{54.55} & 48.21 & \textbf{45.82} & 42.83 & 76.33 & \textbf{76.50} & \textbf{58.90} & 55.85 \\
& Llama-3.1-8B & \textbf{64.03} & 63.43 & \textbf{52.56} & 52.30 & 79.38 & \textbf{79.65} & \textbf{65.32} & 65.13 \\
& Qwen3-0.6B   & \textbf{46.98} & 45.22 & \textbf{33.02} & 31.74 & \textbf{67.63} & 65.89 & \textbf{49.21} & 47.62 \\
& Qwen3-1.7B   & \textbf{59.3}  & 56.96 & \textbf{42.15} & 41.21 & \textbf{71.76} & 71.11 & \textbf{57.74} & 56.43 \\
& Qwen3-4B     & \textbf{69.25} & 68.99 & \textbf{53.58} & 53.33 & 74.54 & \textbf{74.70} & \textbf{65.79} & 65.67 \\
& Qwen3-8B     & 74.46 & \textbf{74.67} & \textbf{57.51} & 56.40 & 77.26 & \textbf{77.80} & \textbf{69.74} & 69.62 \\
& Qwen3-14B    & \textbf{78.69} & 78.54 & \textbf{62.12} & 60.07 & 79.54 & \textbf{80.03} & \textbf{73.45} & 72.88 \\

\midrule

\multirow{8}{*}{\textbf{4-bit}}
& Llama-3.2-1B & 24.12 & \textbf{24.78} & 26.79 & \textbf{30.97} & 50.60 & \textbf{66.70} & 33.84 & \textbf{40.82} \\
& llama-3.2-3B & 23.38 & \textbf{29.72} & 26.02 & \textbf{35.58} & 50.87 & \textbf{71.44} & 33.42 & \textbf{45.58} \\
& llama-3.1-8B & 25.22 & \textbf{45.42} & 25.43 & \textbf{39.42} & 51.52 & \textbf{74.59} & 34.06 & \textbf{53.14} \\
& Qwen3-0.6B   & 24.65 & \textbf{34.02} & 26.19 & \textbf{27.56} & 52.18 & \textbf{62.19} & 34.34 & \textbf{41.26} \\
& Qwen3-1.7B   & 24.33 & \textbf{47.83} & 26.19 & \textbf{32.42} & 50.76 & \textbf{66.05} & 33.76 & \textbf{48.77} \\
& Qwen3-4B     & 24.70 & \textbf{60.09} & 28.07 & \textbf{42.83} & 50.22 & \textbf{72.80} & 34.33 & \textbf{58.57} \\
& Qwen3-8B     & 25.06 & \textbf{66.02} & 27.05 & \textbf{44.88} & 50.27 & \textbf{74.86} & 34.13 & \textbf{61.92} \\
& Qwen3-14B    & 24.68 & \textbf{72.97} & 24.83 & \textbf{49.15} & 49.29 & \textbf{77.75} & 32.93 & \textbf{66.62} \\

\bottomrule
\end{tabularx}
\end{table*}

\begin{table*}[h]
\centering
\setlength{\tabcolsep}{4pt}
\caption{Perplexity ($\downarrow$) on Wikitext-2, PennTreebank, and C4. We compare asymmetric linear quantization vs.\ our Log$_\text{b}$Quant approach. Models L refer to Llama-3.1/3.2, models Q to Qwen3.}
\label{tab:perplexity}
\scriptsize
\begin{tabular}{l!{\hspace{2pt}\vrule width 0.8pt\hspace{2pt}}l*{8}{c}}

\toprule

\multicolumn{2}{c}{\multirow{3}{*}{Model}} &
\multicolumn{2}{c}{WT2} &
\multicolumn{2}{c}{PTB} &
\multicolumn{2}{c}{C4} &
\multicolumn{2}{c}{Avg.} \\

\cmidrule(lr){3-4} \cmidrule(lr){5-6} \cmidrule(lr){7-8} \cmidrule(lr){9-10}

\multicolumn{2}{c}{} &
lin & log & lin & log & lin & log & lin & log \\
\midrule

\multirow{8}{*}{\textbf{8-bit}}
& L-1B   & \textbf{10.49}  & 15.13 & \textbf{17.72}  & 22.00  & \textbf{14.95}  & 24.11 & \textbf{14.39}  & 20.41 \\
& L-3B   & \textbf{8.32}   & 9.46  & \textbf{13.35}  & 15.70  & \textbf{12.07}  & 14.38 & \textbf{11.25}  & 13.18 \\
& L-8B   & \textbf{6.88}   & 6.97  & \textbf{11.13}  & 11.40  & \textbf{10.40}  & 10.60 & \textbf{9.47}   & 9.66 \\
& Q-0.6B & \textbf{22.60}  & 24.10 & \textbf{43.83}  & 47.53  & \textbf{30.69}  & 32.25 & \textbf{32.37}  & 34.63 \\
& Q-1.7B & \textbf{18.09}  & 20.31 & \textbf{32.30}  & 36.55  & \textbf{23.04}  & 25.11 & \textbf{24.48}  & 27.32 \\
& Q-4B   & \textbf{14.69}  & 15.21 & \textbf{24.49}  & 25.77  & \textbf{20.27}  & 21.10 & \textbf{19.82}  & 20.69 \\
& Q-8B   & \textbf{10.41}  & 10.57 & \textbf{17.17}  & 17.50  & \textbf{15.76}  & 15.87 & \textbf{14.45}  & 14.65 \\
& Q-14B  & \textbf{9.26}   & 9.27  & \textbf{15.40}  & 15.60  & \textbf{14.20}  & 14.10 & \textbf{12.95}  & 12.99 \\

\midrule

\multirow{8}{*}{\textbf{4-bit}}
& L-1B   & \num{1.88e4} & \textbf{18.26} & \num{4.09e4} & \textbf{32.81} & \num{3.88e4} & \textbf{30.78} & \num{3.28e4} & \textbf{27.28} \\
& L-3B   & \num{9.63e4} & \textbf{13.33} & \num{1.38e5} & \textbf{24.61} & \num{7.46e4} & \textbf{21.46} & \num{1.03e5} & \textbf{19.80} \\
& L-8B   & \num{1.53e5} & \textbf{10.30} & \num{1.60e5} & \textbf{18.76} & \num{1.17e5} & \textbf{17.98} & \num{1.44e5} & \textbf{15.68} \\
& Q-0.6B & \num{5.37e3} & \textbf{31.85} & \num{3.86e4} & \textbf{65.23} & \num{1.47e4} & \textbf{45.31} & \num{1.95e4} & \textbf{47.46} \\
& Q-1.7B & \num{2.76e4} & \textbf{22.10} & \num{8.12e4} & \textbf{43.46} & \num{3.16e4} & \textbf{29.23} & \num{4.68e4} & \textbf{31.60} \\
& Q-4B   & \num{1.63e3} & \textbf{16.94} & \num{4.29e3} & \textbf{28.95} & \num{6.72e3} & \textbf{23.90} & \num{4.21e3} & \textbf{23.26} \\
& Q-8B   & 325.18       & \textbf{11.82} & \num{1.84e3} & \textbf{19.63} & \num{2.11e3} & \textbf{18.00} & \num{1.42e3} & \textbf{16.48} \\
& Q-14B  & 241.30       & \textbf{10.65} & \num{1.34e3} & \textbf{18.94} & \num{1.44e3} & \textbf{16.28} & \num{1.01e3} & \textbf{15.29} \\

\bottomrule
\end{tabular}
\end{table*}

Language modeling performance is measured in terms of perplexity on the validation splits of WikiText-2~\cite{merity2016pointersentinelmixturemodels}, Penn Treebank (PTB)~\cite{marcus1993building} and 1000 samples of the C4 dataset~\cite{Raffel2020T5}.
Perplexity is calculated as an average of blocks of length 2048 without sliding window.
The results can be found in Table~\ref{tab:perplexity}.
Performance of the unquantized models can be found in Table~\ref{tab:perplexity_base} (Appendix~\ref{ap:tabs}).

For 8-bit quantization, the linear quantization approach consistently outperforms our own approach in terms of perplexity, even though the differences between the two approaches become marginal with increasing model size.
For our downstream evaluation, linear still performs better for ARC-C and most of MMLU, however, differences between both approaches are low again.
Both approaches tie on the PIQA dataset.
However, it is worth noting that both approaches almost saturate the original 16-bit models' performance.
Further, we conducted additional experiments using an equally sized sample of the RedPajama~\cite{weber_redpajama_2024} dataset for calibration on the Llama models.
In those runs, our logarithmic approach outperformed asymmetric linear quantization across almost benchmarks (albeit also marginally) and average scores for both approaches were consistently slightly better (see Tables~\ref{tab:accuracy_alt} and \ref{tab:perplexity_alt} in Appendix~\ref{ap:tabs}.
For ensuring a fair comparison and aligning our methodology with the literature, we decided to report the scores after calibrating on WikiText-2 instead.

The more interesting effects become visible when quantizing to 4-bit representations.
In our downstream benchmarks, linear quantization completely breaks the models' reasoning capabilities, with results almost identical to random guessing.
On the other hand, our own approach still performs well compared to the baseline even though performance slightly deteriorates.
The perplexity results further emphasize this pattern: The linearly quantized models can no longer properly predict next tokens accurately.
Performance is slightly recovered with increasing model size, but perplexity is still significantly higher than log-quantized models.

The experiments also show that our values of $\varepsilon$ for the energy-based pruning prove robust and generalize well over different models.
While more lenient pruning at 8-bit precision is sufficient, weights of low magnitude need to be removed more aggressively for 4-bit quantization as a tighter effective weight range improves approximation for small codebooks.

\subsection{Efficiency Benchmarks}
To measure the efficiency of our approach, we compare the model throughput (see Table~\ref{tab:tps}) and peak memory usage (see Table~\ref{tab:memory}) of our approach at 4-bit compared to the baseline models (unquantized, bfloat16).
In all experiments in this section, we produce exactly 128 new tokens per input prompt (input prompts being identical in all batch samples) with greedy search.
Runs are repeated until wallclock times statistically stabilize, and are eventually averaged.

It is important to note that, in contrast to memory savings, speed is not the focus of our approach since dequantization at inference time introduces an overhead.
\begin{table*}[h]
\centering
\setlength{\tabcolsep}{4pt}

\caption{Model throughput in tokens per second ($\uparrow$). We compare the original 16-bit bfloat model (full) against our 4-bit Log$_\text{b}$Quant model (log) for 3 different batch sizes. Models L refer to Llama-3.1/3.2, models Q to Qwen3.}
\label{tab:tps}
\scriptsize
\begin{tabularx}{\textwidth}{l*{10}{>{\centering\arraybackslash}X}}
\toprule

\multirow{3}{*}{Model} &
\multicolumn{3}{c}{batch size 1} &
\multicolumn{3}{c}{batch size 32} &
\multicolumn{3}{c}{batch size 128} \\

\cmidrule(lr){2-4}
\cmidrule(lr){5-7}
\cmidrule(lr){8-10}

&
full & log & speedup &
full & log & speedup &
full & log & speedup\\
\midrule

L-1B   & 78.06  & 81.74  & $1.05\times$  & 2435.39  & 2685.20  & $1.10\times$  & 9743.38  & 8225.03  & $0.84\times$ \\
L-3B   & 45.51  & 49.18  & $1.08\times$  & 1430.64  & 1565.21  & $1.09\times$  & 4673.76  & 3659.98  & $0.78\times$ \\
L-8B   & 31.23  & 40.88  & $1.31\times$  &  875.01  & 1231.72  & $1.41\times$  & 2751.18  & 1704.08  & $0.62\times$ \\
Q-0.6B & 39.59  & 39.83  & $1.01\times$  & 1225.51  & 1279.95  & $1.04\times$  & 4408.56  & 4965.88  & $1.13\times$ \\
Q-1.7B & 36.16  & 38.74  & $1.07\times$  & 1157.96  & 1218.55  & $1.05\times$  & 4521.08  & 4603.01  & $1.02\times$ \\
Q-4B   & 27.68  & 30.11  & $1.09\times$  &  904.29  &  967.06  & $1.07\times$  & 3364.21  & 2527.99  & $0.75\times$ \\
Q-8B   & 27.61  & 28.33  & $1.03\times$  &  785.01  &  885.33  & $1.13\times$  & 2478.66  & 1572.74  & $0.63\times$ \\
Q-14B  & 17.42  & 26.24  & $1.51\times$  &  494.13  &  723.44  & $1.46\times$  & 1614.11  &  909.88  & $0.56\times$ \\

\bottomrule
\end{tabularx}
\end{table*}
However, we nevertheless achieve minor to moderate speedups, in particular with growing model size.
Similar results are observed by~\cite{dettmers2022llmint88bitmatrixmultiplication}, whose approach we followed for the linear implementation.
We attribute the throughput increase to the aforementioned memory wall: As weights are streamed in a compressed representation, we are no longer bound by memory bandwidth.
Speedups diminish or turn into slowdowns when running at a batch size of 128.
We assume this is due to excessive reads of the shared memory banks for dequantizing using LUTs.
This assumption is further supported by comparing Qwen3-0.6B/1.7B against Llama-3.2-1B:
While the Llama model has only 16 hidden layers, both Qwen3 models have 28 instead, implying a lower number of weights per layer and thus fewer simultaneous reads.
\begin{table*}[h]
\centering
\setlength{\tabcolsep}{4pt}

\caption{Model peak memory usage in GB ($\downarrow$). We compare the original 16-bit bfloat model (full) against our 4-bit Log$_\text{b}$Quant model (log) for 3 different batch sizes. Models L refer to Llama-3.1/3.2, models Q to Qwen3.}
\label{tab:memory}
\scriptsize
\begin{tabularx}{\textwidth}{l*{10}{>{\centering\arraybackslash}X}}
\toprule

\multirow{3}{*}{Model} &
\multicolumn{3}{c}{batch size 1} &
\multicolumn{3}{c}{batch size 32} &
\multicolumn{3}{c}{batch size 128} \\

\cmidrule(lr){2-4}
\cmidrule(lr){5-7}
\cmidrule(lr){8-10}

&
full & log & savings &
full & log & savings &
full & log & savings\\
\midrule

L-1B   &  2.43  &  1.20  & 50.40\%  &  2.60  &  1.37  & 47.09\%  &  3.13  &  1.90  & 39.12\%\\
L-3B   &  6.30  &  2.43  & 61.44\%  &  6.80  &  2.93  & 56.92\%  &  8.35  &  4.48  & 46.35\%\\
L-8B   & 15.71  &  5.87  & 62.62\%  & 16.28  &  6.44  & 60.44\%  & 18.03  &  8.19  & 54.56\%\\
Q-0.6B &  1.19  &  0.73  & 38.64\%  &  1.69  &  1.23  & 27.13\%  &  3.26  &  2.80  & 14.11\%\\
Q-1.7B &  3.39  &  1.54  & 54.62\%  &  3.89  &  2.04  & 47.54\%  &  5.45  &  3.60  & 33.92\%\\
Q-4B   &  7.89  &  2.85  & 63.90\%  &  8.52  &  3.48  & 59.14\%  & 10.49  &  5.45  & 48.04\%\\
Q-8B   & 16.03  &  6.24  & 61.09\%  & 16.66  &  6.87  & 58.76\%  & 16.66  &  8.84  & 52.55\%\\
Q-14B  & 28.88  & 10.10  & 65.03\%  & 29.58  & 10.80  & 63.49\%  & 31.75  & 12.97  & 59.15\%\\

\bottomrule
\end{tabularx}
\end{table*}

Our main focus is memory efficiency, at which Log$_\text{b}$Quant excels.
Since we provide a weight-only quantization approach, memory savings decrease with batch size as more memory is required for activations and KV cache.
As the unquantized embedding layers and language modeling heads contribute negatively to memory usage, savings deviate from the optimal 75\% possible with full 4-bit quantization, with relatively higher values for increasing model sizes.
\section{Discussion and Limitations}
Our novel quantization scheme, Log$_\text{b}$Quant, shows superior benchmark performance against asymmetric linear quantization at 4-bit and comes with minor to moderate speedups and significant memory savings compared to the unquantized model.
In particular, it is worth noting that Qwen3-14B, quantized with our method, shows an increase in token throughput of $1.51\times$ with a $65\%$ memory reduction to 10.10 GB with an average accuracy reduction of about 6 percent points, making it the best model (from linear, logarithmic and no quantization) in our experiments that can be run on common consumer-grade graphics cards with 12 GB VRAM, such as the Nvidia RTX 5070.
Log$_\text{b}$Quant performs well for small codebooks since high magnitude outlier values can be appropriately represented without skewing precision in high-density regions of the weight distributions, unlike asymmetric linear approaches.

Our results show that coarse, tensor-level quantization does not need to imply significant loss of performance at low bitwidths, but instead can show similarly good results as less strong compression.
However, logarithmic quantization at 8-bit comes with no advantage over linear approaches, where both methods almost saturate or exceed base model performance.
Since the overhead of direct dequantization or significantly larger LUTs imposes a stronger constraint than linear quantization to int8, the latter approach is preferable in this scenario.
Further, quantization to the FP8 format, which becomes more broadly available on accelerators these days~\cite{micikevicius2022fp8}, may resolve issues of both linear and Log$_\text{b}$Quant int8 approaches in this scenario.

The speedup of Log$_\text{b}$Quant is further theoretically limited by the free choice of bases $b$ in weight-and-activation quantization:
Since bases differ between tensors and are real-valued, the approximations of multiplication and particularly of addition become unstable when done directly in int8.
This stands in contrast to weight-and-activation linear quantization schemes, which can achieve high throughput by directly calculating in int8.
\section{Conclusion and Future Work}
We introduce Log$_\text{b}$Quant, a novel quantization scheme, and show that performs particularly well in several benchmarks at coarse granularity settings and low bitwidth precision against asymmetric quantization.
Additionally to tangible speedups, which were not the main focus of this work, significant memory savings allow us to quantize models with 14B parameters to fit into 12 GB of VRAM at inference time with low impact on performance, which is crucial for acceleration on current consumer hardware.

While our results already showcase the benefit of our method for the tested settings, generalization to even lower bitwidth representations are of particular interest.
As our primary goal was to reduce memory requirements for consumer and edge devices, activation quantization was out-of-scope as memory savings for personal use are negligible.
Nevertheless, extending our approach to activations and KV cache could still be of interest.
Alternatively, testing Log$_\text{b}$Quant in combination with other methods orthogonal to ours, such as TurboQuant~\cite{zandieh2025turboquant}, may improve memory efficiency even further.
 \bibliographystyle{splncs04}
 \bibliography{refs}
%Author, F.: Article title. Journal \textbf{2}(5), 99--110 (2016)
%Author, F., Author, S.: Title of a proceedings paper. In: Editor,
%F., Editor, S. (eds.) CONFERENCE 2016, LNCS, vol. 9999, pp. 1--13.
%Springer, Heidelberg (2016). \doi{10.10007/1234567890}
%Author, F., Author, S., Author, T.: Book title. 2nd edn. Publisher,
%Author, A.-B.: Contribution title. In: 9th International Proceedings
%on Proceedings, pp. 1--2. Publisher, Location (2010)
%LNCS Homepage, \url{http://www.springer.com/lncs}. Last accessed 4
%\end{thebibliography}
    \appendix
    \section{Additional Tables}\label{ap:tabs}
    \begin{table*}[h]
\centering
\setlength{\tabcolsep}{4pt}

\caption{Accuracy ($\uparrow$) on MMLU and normalized accuracy ($\uparrow$) on ARC-C and PIQA, respectively. Reported values are for the base models in BF16.}
\label{tab:accuracy_base}
\scriptsize
\begin{tabularx}{\textwidth}{l*{5}{>{\centering\arraybackslash}X}}
\toprule

Model &
MMLU &
ARC-C &
PIQA &
Avg. \\
\midrule

Llama-3.2-1B & 37.84 & 37.03 & 74.65 & 49.84 \\
Llama-3.2-3B & 55.01 & 46.42 & 77.80 & 59.74 \\
Llama-3.1-8B & 63.87 & 54.86 & 81.01 & 66.58 \\
Qwen3-0.6B   & 40.09 & 33.96 & 67.63 & 47.23 \\
Qwen3-1.7B   & 55.55 & 43.09 & 72.20 & 56.95 \\
Qwen3-4B     & 68.37 & 54.01 & 75.19 & 65.86 \\
Qwen3-8B     & 72.90 & 56.48 & 77.69 & 69.11 \\
Qwen3-14B    & 77.15 & 60.15 & 79.76 & 72.35 \\

\bottomrule
\end{tabularx}
\end{table*}

\begin{table*}[h]
\centering
\setlength{\tabcolsep}{4pt}
\caption{Perplexity ($\downarrow$) on Wikitext-2, PennTreebank, and C4. Reported values are for the base models in BF16.}
\label{tab:perplexity_base}
\scriptsize
\begin{tabularx}{\textwidth}{l*{5}{>{\centering\arraybackslash}X}}
\toprule

Model &
WT2 &
PTB &
C4 &
Avg. \\
\midrule

Llama-3.2-1B & 10.05 & 16.78 & 13.87 & 13.57 \\
Llama-3.2-3B &  7.97 & 12.81 & 11.25 & 10.68 \\
Llama-3.1-8B &  6.52 & 10.74  & 9.52 &  8.93 \\
Qwen3-0.6B   & 22.44 & 43.08 & 30.25 & 31.92 \\
Qwen3-1.7B   & 18.00 & 32.48 & 22.83 & 24.44 \\
Qwen3-4B     & 14.64 & 24.52 & 20.32 & 19.83 \\
Qwen3-8B     & 10.32 & 16.97 & 15.56 & 14.28 \\
Qwen3-14B    &  9.16 & 14.93 & 13.92 & 12.67 \\

\bottomrule
\end{tabularx}
\end{table*}

\begin{table*}[h]
\centering
\setlength{\tabcolsep}{4pt}

\caption{Accuracy ($\uparrow$) on MMLU and normalized accuracy ($\uparrow$) on ARC-C and PIQA, respectively, after calibrating on RedPajama  for 8-bit quantization. We compare asymmetric linear quantization vs.\ our Log$_\text{b}$Quant approach.}
\label{tab:accuracy_alt}
\scriptsize
\begin{tabularx}{\textwidth}{l*{9}{>{\centering\arraybackslash}X}}
\toprule

\multicolumn{1}{c}{\multirow{3}{*}{Model}} &
\multicolumn{2}{c}{MMLU} &
\multicolumn{2}{c}{ARC-C} &
\multicolumn{2}{c}{PIQA} &
\multicolumn{2}{c}{Avg.} \\

\cmidrule(lr){2-3}
\cmidrule(lr){4-5}
\cmidrule(lr){6-7}
\cmidrule(lr){8-9}

&
lin & log
& lin & log
& lin & log
& lin & log \\
\midrule

Llama-3.2-1B & 34.80 & \textbf{37.67} & 36.01 & \textbf{37.12} & \textbf{74.86} & 74.54 & 48.56 & \textbf{49.78} \\
Llama-3.2-3B & 54.46 & \textbf{54.94} & \textbf{45.99} & 45.73 & 77.37 & \textbf{77.91} & 59.27 & \textbf{59.53} \\
Llama-3.1-8B & 62.29 & \textbf{64.11} & 53.24 & \textbf{54.52} & 80.52 & \textbf{80.79} & 65.35 & \textbf{66.47} \\
\bottomrule
\end{tabularx}
\end{table*}

\begin{table*}[h]
\centering
\setlength{\tabcolsep}{4pt}
\caption{Perplexity ($\downarrow$) on Wikitext-2, PennTreebank, and C4 after calibrating on RedPajama for 8-bit quantization. We compare asymmetric linear quantization vs.\ our Log$_\text{b}$Quant approach. L refers to Llama-3.1/3.2.}
\label{tab:perplexity_alt}
\scriptsize
\begin{tabularx}{\textwidth}{l*{9}{>{\centering\arraybackslash}X}}

\toprule

\multicolumn{1}{c}{\multirow{3}{*}{Model}} &
\multicolumn{2}{c}{WT2} &
\multicolumn{2}{c}{PTB} &
\multicolumn{2}{c}{C4} &
\multicolumn{2}{c}{Avg.} \\

\cmidrule(lr){2-3} \cmidrule(lr){4-5} \cmidrule(lr){6-7} \cmidrule(lr){8-9}

&
lin & log & lin & log & lin & log & lin & log \\
\midrule
Llama-3.2-1B   & 10.33  & \textbf{10.12} & 17.14  & \textbf{16.90}  & 14.31  & \textbf{13.95} & 13.93  & \textbf{13.66} \\
Llama-3.2-3B   & 8.08   & \textbf{8.00}  & 12.97  & \textbf{12.86}  & 11.45  & \textbf{11.29} & 10.83  & \textbf{10.72} \\
Llama-3.2-8B   & 6.74   & \textbf{6.55}  & 10.97  & \textbf{10.78}  & 9.98   & \textbf{9.58}  & 9.23   & \textbf{8.97} \\
\bottomrule
\end{tabularx}
\end{table*}
\end{document}